\documentclass[a4paper,UKenglish,cleveref, autoref, thm-restate]{lipics-v2019}

\newcommand{\Baire}{\mathbb{N}^\mathbb{N}}
\newcommand{\dom}{\operatorname{dom}}
\newcommand{\sat}{\uparrow}
\newcommand{\id}{\textnormal{id}}

\usepackage{ulem}

\title{A Computability Perspective on (Verified) Machine Learning}

\titlerunning{A Computability Perspective on (Verified) Machine Learning} 

\author{Tonicha Crook}{Department of Computer Science, Swansea University, UK}{t.m.crook15@outlook.com}{https://orcid.org/0000-0002-4882-9999}{This author is supported by the UKRI AIMLAC CDT, cdt-aimlac.org, grant no. EP/S023992/1.}

\author{Jay Morgan}{Department of Computer Science, Swansea University, UK}{j.p.morgan@swansea.ac.uk}{https://orcid.org/0000-0003-3719-362X}{}

\author{Arno Pauly}{Department of Computer Science, Swansea University, UK \and \url{https://www.cs.swan.ac.uk/~cspauly/}}{arno.m.pauly@gmail.com}{https://orcid.org/0000-0002-0173-3295}{}

\author{Markus Roggenbach}{Department of Computer Science, Swansea University, UK}{m.roggenbach@swansea.ac.uk}{https://orcid.org/0000-0002-3819-2787}{}

\authorrunning{T.~Crook, J.~Morgan, A.~Pauly \& M.~Roggenbach} 

\Copyright{Tonicha Crook, Jay Morgan, Arno Pauly and Markus Roggenbach} 



\category{} 

\relatedversion{} 

\supplement{}

\nolinenumbers 

\hideLIPIcs  

\EventEditors{John Q. Open and Joan R. Access}
\EventNoEds{2}
\EventLongTitle{42nd Conference on Very Important Topics (CVIT 2016)}
\EventShortTitle{CVIT 2016}
\EventAcronym{CVIT}
\EventYear{2016}
\EventDate{December 24--27, 2016}
\EventLocation{Little Whinging, United Kingdom}
\EventLogo{}
\SeriesVolume{42}
\ArticleNo{23}

\begin{document}

\maketitle

\begin{abstract}
There is a strong consensus that combining the versatility of machine learning with the assurances given by formal verification is highly desirable. It is much less clear what verified machine learning should mean exactly. We consider this question from the (unexpected?) perspective of computable analysis. This allows us to define the computational tasks underlying verified ML in a model-agnostic way, and show that they are in principle computable.
\end{abstract}

\section{Introduction}

Machine Learning (ML) concerns the process of building both predictive and generative models through the use of optimisation procedures. The remarkable success of ML methods in various domains raises the question of how much trust one can put into the responses that a ML model provides. As ML models are also applied in critical domains, some form of verification might be adequate. 

However, due to the widespread use of non-discrete mathematics in ML, traditional verification techniques are hard to apply to its artefacts. Furthermore, many ML applications lack specifications in form of, say,  an input/output relationship, on which many `classical' verification approaches are based. A typical example of this would be an ML application that shall decide if a given picture depicts a cat. Lacking a specification, what kind of properties can be verified?

Using the toolset of computable analysis, in this paper we present an investigation into what kind of verification questions are answerable in principle about ML models -- irrespective of the particular ML framework applied. 


For the purpose of this paper, we will focus upon the notion of a ML classifier. A classifier
takes as input some description of an object, say in 
form of a vector of real numbers. As an output a classifier either provides a
colour, which we take to be an integer from the set $\mathbf{k} = \{0,\ldots,k-1\}, k > 0$, or it does not give an answer, i.e., we consider classifiers to be semi-decision procedures. 
%
%
Simple examples of properties that such a classifier is $f$ might exhibit, include: 
\begin{itemize}
    \item Can $f$ output a specific colour for points stemming from a given region?
    \item Is $f$ constant on a given region?
    \item Are there two `close' points that $f$ maps to distinct colours?
\end{itemize}
We will show for these and other properties that they are indeed computable (under some `natural' assumptions). 

One specific verification task that has caught the attention of the ML community is to find \emph{adversarial examples} \cite{szegedy2013intriguing,Goodfellow-et-al-2016,Huang2016} or to prevent them from occurring. One says that an adversarial example occurs when a `small' change to the input results in a `large' change in output or a miss-classifications is made by the predictive model (i.e.~our third task above). For example, given a correctly classified image, small, even unnoticeable changes to the pixels in the image can vastly change the classification output. The existence of adversarial examples can be catastrophic in the context of fully-autonomous vehicles that use sensor-captured data as input for Deep Neural Network (DNN) models. In such situations, \cite{Huang2016} explains how lighting conditions and angles, as well as defects in sensor equipment themselves, may contain many realistic adversarial examples. We show that certain variations of this task are computable without the need to refer to specifics of an ML framework.


Though we will present semi-decision procedures in this paper, our algorithms are not meant for implementation. For our foundational work on what kind of verification is possible in principle, it is beneficial to work in a framework which is agnostic to ML methods. However, when it comes to the task of actually developing `useful' verification algorithms, it will be essential to take into account the specifics of the method used.

While the language of computable analysis was a natural choice for us, we do not claim that this is the only way of how the challenge of `verified ML' can be addressed. Providing an abstract perspective will, we hope, help further researchers in the field to develop new ideas and communicate more efficiently.

The rest of the paper is organised as follows: in Section \ref{sec:background} we explore the concepts required for this paper within computable analysis. Classification tasks are considered within Section \ref{sec:AdvEx}, where we look into adversarial examples. We explore different techniques and how well they manage adversarial examples/attacks. The training of classifiers is discussed in Section \ref{sec:robust_learners}, where we explore the robustness of learners and sparsity of data. Finally, in Section \ref{sec:whatspaces} we examine different demands we could hold over our spaces, such as Hausdorffness and compactness.

\section{Computing with real numbers and other non-discrete data types}
\label{sec:background}

The mathematics behind the most popular notions of ML is  based on real numbers; whether it is on a fundamental level such as the use of Sigmoid activation functions in DNNs, the vector space structure for Support Vector Machine (SVM), or more abstract ideas such as the data manifold hypothesis \cite{narayanan2010sample}. While the actual implementations will use floating point numbers rather than exact real arithmetic, this still suggests that computable analysis is a reasonable framework for studying algorithmic issues in an abstract setting.

Computable analysis as a field is not just concerned with computability of operations on the real numbers, but far more generally with non-necessarily discrete domains. One question we strive to answer here is what properties of the domain are actually needed (or at least useful) to obtain the fundamental results by keeping track of what we are using where. Working with effectively locally-compact computable metric spaces turns out to be a promising candidate, as we will argue in Section \ref{sec:whatspaces}.

Standard sources for computable analysis include the textbook \cite{weihrauchd} by Weihrauch, around which the modern field coalesced; as well as the tutorial article \cite{brattkaintro}. The approach we follow is the one outlined in \cite{pauly-synthetic}, to which we refer to reader for details.

The basic structure in computable analysis is a \emph{represented space}, which carries a notion of computability via a coding of elements by infinite binary sequences. Core aspects relevant for us include the function space construction $\mathcal{C}(\mathbf{X},\mathbf{Y})$, where a function from $\mathbf{X}$ to $\mathbf{Y}$ is coded by just enough information to evaluate it. We refer to several hyperspaces of subsets, such as the open sets $\mathcal{O}(\mathbf{X})$ with just enough information to recognise membership (given an $U \in \mathcal{O}(\mathbf{X})$ and $x \in X$, the property $x\in U$ is semidecidable), compact subsets $\mathcal{K}(\mathbf{X})$ with just enough information to recognise containment in open sets (given $A \in \mathcal{K}(\mathbf{X})$ and $U \in \mathcal{O}(\mathbf{X})$, the property $A \subseteq U$ is semidedicable), and overt subsets $\mathcal{V}(\mathbf{X})$ with just enough information to recognise intersecting open sets (given $A \in \mathcal{V}(\mathbf{X})$ and $U \in \mathcal{O}(\mathbf{X})$, the property $A \cap U \neq \emptyset$ is semidedicable). Knowing $A \in (\mathcal{V} \wedge \mathcal{K})(\mathbf{X})$ means knowing $A \in \mathcal{V}(\mathbf{X})$ and $A \in \mathcal{K}(\mathbf{X})$ at the same time.

The spaces $\mathbb{R}^n$ can be considered as represented spaces by encoding real vectors as limits of sequences of rational vectors converging with speed $2^{-n}$. This in particular ensures that the maps $(x,r) \mapsto \overline{B}(x,r) : \mathbb{R}^n \times \mathbb{R} \to \mathcal{K}(\mathbf{\mathbb{R}^n})$ and $(x,r) \mapsto \overline{B}(x,r) : \mathbb{R}^n \times \mathbb{R} \to \mathcal{V}(\mathbf{\mathbb{R}^n})$ and computable, where $\overline{B}(x,r) = \{y \in \mathbb{R}^n \mid d(x,y) \leq r\}$ are the closed balls (with $d$ being any `reasonable' distance, e.g.~the Euclidean).

An important facet of computable analysis is that equality is typically not decidable; in particular it is not for the spaces $\mathbb{R}^n$. For these, inequality is semidecidable though (which makes them computably Hausdorff). If equality is also semidecidable, a space is called computably discrete. 

\subsection*{Some mathematical detail}
The following summarizes the formal definitions and key properties of the most important notions for our paper. It is taken near verbatim from \cite{paulydebrecht3-csl}. As mentioned above, a more comprehensive treatment is found in \cite{pauly-synthetic}.

\begin{definition}
A represented space is a pair $\mathbf{X} = (X, \delta_\mathbf{X})$ where $X$ is a set and $\delta_\mathbf{X} : \subseteq \Baire \to X$ is a partial surjection (the notation $:\subseteq$ denotes partial functions). A function between represented spaces is a function between the underlying sets.
\end{definition}

\begin{definition}
For $f : \subseteq \mathbf{X} \to \mathbf{Y}$ and $F : \subseteq \Baire \to \Baire$, we call $F$ a realizer of $f$ (notation $F \vdash f$), iff $\delta_Y(F(p)) = f(\delta_X(p))$ for all $p \in \dom(f\delta_X)$. A map between represented spaces is called computable (continuous), iff it has a computable (continuous) realizer.
\end{definition}

Two represented spaces of particular importance are the integers $\mathbb{N}$ and Sierpi\'nski space $\mathbb{S}$. The represented space $\mathbb{N}$ has as underlying set $\mathbb{N}$ and the representation $\delta_\mathbb{N} : \Baire \to \mathbb{N}$ defined by $\delta_\mathbb{N}(p) = p(0)$. The Sierpi\'nski space $\mathbb{S}$ has the underlying set $\{\top,\bot\}$ and the representation $\delta_\mathbb{S}$ with $\delta_\mathbb{S}(0^\omega) = \bot$ and $\delta_\mathbb{S}(p) = \top$ for $p \neq 0^\omega$.

Represented spaces have binary products, defined in the obvious way: The underlying set of $\mathbf{X} \times \mathbf{Y}$ is $X \times Y$, with the representation $\delta_{\mathbf{X} \times \mathbf{Y}}(\langle p, q\rangle) = (\delta_\mathbf{X}(p),\delta_\mathbf{Y}(q))$. Here $\langle \ , \ \rangle : \Baire \times \Baire \to \Baire$ is the pairing function defined via $\langle p, q\rangle(2n) = p(n)$ and $\langle p, q\rangle(2n+1) = q(n)$.

A central reason why the category of represented spaces is such a convenient setting lies in the fact that it is cartesian closed. We have available a function space construction $\mathcal{C}(\cdot, \cdot)$, where the represented space $\mathcal{C}(\mathbf{X},\mathbf{Y})$ has as underlying set the continuous functions from $\mathbf{X}$ to $\mathbf{Y}$, represented in such a way that the evaluation map $(f, x) : \mathcal{C}(\mathbf{X},\mathbf{Y}) \times \mathbf{X} \to \mathbf{Y}$ becomes computable. This can be achieved, e.g., by letting $nq$ represent $f$, if the $n$-th Turing machine equipped with oracle $q$ computes a realizer of $f$. This also makes currying, uncurrying and composition all computable maps.

Having available to us the space $\mathbb{S}$ and the function space construction, we can introduce the spaces $\mathcal{O}(\mathbf{X})$ and $\mathcal{A}(\mathbf{X})$ of open and closed subsets respectively of a given represented space $\mathbf{X}$. For this, we identify an open subset $U$ of $\mathbf{X}$ with its (continuous) characteristic function $\chi_U : \mathbf{X} \to \mathbb{S}$, and a closed subset with the characteristic function of the complement. As countable join (or) and binary meet (and) on $\mathbb{S}$ are computable, we can conclude that open sets are uniformly closed under countable unions, binary intersections and preimages under continuous functions by merely using elementary arguments about function spaces. The space $\mathcal{A}(\mathbf{X})$ corresponds to the upper Fell topology \cite{fell} on the hyperspace of closed sets.

Note that neither negation $\mathalpha{\neg} : \mathbb{S} \to \mathbb{S}$ (i.e.~mapping $\top$ to $\bot$ and $\bot$ to $\top$) nor countable meet (and) $\bigwedge : \mathcal{C}(\mathbb{N},\mathbb{S}) \to \mathbb{S}$ (i.e.~mapping the constant sequence $(\top)_{n \in \mathbb{N}}$ to $\top$ and every other sequence to $\bot$) are continuous or computable operations.

We need two further hyperspaces, which both will be introduced as subspaces of $\mathcal{O}(\mathcal{O}(\mathbf{X}))$. The space $\mathcal{K}(\mathbf{X})$ of saturated compact sets identifies $A \subseteq \mathbf{X}$ with $\{U \in \mathcal{O}(\mathbf{X}) \mid A \subseteq U\} \in \mathcal{O}(\mathcal{O}(\mathbf{X}))$. Recall that a set is saturated, iff it is equal to the intersection of all open sets containing it (this makes the identification work). The saturation of $A$ is denoted by $\sat{A} := \bigcap \{U \in \mathcal{O}(\mathbf{X}) \mid A \subseteq U\}$. Compactness of $A$ corresponds to $\{U \in \mathcal{O}(\mathbf{X}) \mid A \subseteq U\}$ being open itself. The dual notion to compactness is \emph{overtness}\footnote{This notion is much less known than compactness, as it is classically trivial. It is crucial in a uniform perspective, though. The term \emph{overt} was coined by Taylor \cite{taylor}, based on the observation that these sets share several closure properties with the open sets.}. We obtain the space $\mathcal{V}(\mathbf{X})$ of overt sets by identifying a closed set $A$ with $\{U \in \mathcal{O}(\mathbf{X}) \mid A \cap U \neq \emptyset\} \in \mathcal{O}(\mathcal{O}(\mathbf{X}))$. The space $\mathcal{V}(\mathbf{X})$ corresponds to the lower Fell (equivalently, the lower Vietoris) topology.

Aligned with the definition of the compact and overt subsets of a space, we can also define when a space itself is compact (respectively overt):

\begin{definition}
A represented space $\mathbf{X}$ is (computably) compact, iff $\textrm{isFull} : \mathcal{O}(\mathbf{X}) \to \mathbb{S}$ mapping $X$ to $\top$ and any other open set to $\bot$ is continuous (computable). Dually, it is (computably) overt, iff $\textrm{isNonEmpty} : \mathcal{O}(\mathbf{X}) \to \mathbb{S}$ mapping $\emptyset$ to $\bot$ and any non-empty open set to $\top$ is continuous (computable).
\end{definition}

The relevance of $\mathcal{K}(\mathbf{X})$ and $\mathcal{V}(\mathbf{X})$ is found in particular in the following characterisations, which show that compactness just makes universal quantification preserve open predicates, and dually, overtness makes existential quantification preserve open predicates.

\begin{proposition}[{\cite[Proposition 40]{pauly-synthetic}}]
\label{prop:exists}
The map $\exists : \mathcal{O}(\mathbf{X} \times \mathbf{Y}) \times \mathcal{V}(\mathbf{X}) \to \mathcal{O}(\mathbf{Y})$ defined by $\exists(R, A) = \{y \in Y \mid \exists x \in A \ (x, y) \in R\}$ is computable. Moreover, whenever $\exists : \mathcal{O}(\mathbf{X} \times \mathbf{Y}) \times \mathcal{S}(\mathbf{X}) \to \mathcal{O}(\mathbf{Y})$ is computable for some hyperspace $\mathcal{S}(\mathbf{X})$ and some space $\mathbf{Y}$ containing a computable element $y_0$, then $\overline{\phantom{A}} : \mathcal{S}(\mathbf{X}) \to \mathcal{V}(\mathbf{X})$ is computable, where $\overline{\phantom{A}}$ denotes topological closure.
\end{proposition}

\begin{proposition}[{\cite[Proposition 42]{pauly-synthetic}}]
\label{prop:forall}
The map $\forall : \mathcal{O}(\mathbf{X} \times \mathbf{Y}) \times \mathcal{K}(\mathbf{X}) \to \mathcal{O}(\mathbf{Y})$ defined by $\forall(R, A) = \{y \in Y \mid \forall x \in A \ (x, y) \in R\}$ is computable. Moreover, whenever $\forall : \mathcal{O}(\mathbf{X} \times \mathbf{Y}) \times \mathcal{S}(\mathbf{X}) \to \mathcal{O}(\mathbf{Y})$ is computable for some hyperspace $\mathcal{S}(\mathbf{X})$ and some space $\mathbf{Y}$ containing a computable element $y_0$, then $\sat{\id} : \mathcal{S}(\mathbf{X}) \to \mathcal{K}(\mathbf{X})$ is computable.
\end{proposition}

\section{Verifying classifiers and adversarial examples} \label{sec:AdvEx}
We consider classification tasks only. This means that a trained model will take as input some description of an object, and either outputs a class (which we take to be an integer from $\mathbf{k} = \{0,\ldots,k-1\}$, $k > 0$), or it does not give an answer. Here, not giving an answer can happen by the algorithm failing to terminate, rather than by an explicit refusal to select a class. This is important to handle connected domains such as the reals, in light of the continuity of all computable functions. Formally, we are dealing with the represented space $\mathbf{k}_{\bot}$ here, which contains the elements $\{0,\ldots,k-1,\bot\}$, where $0^\omega$ is the only name for $\bot$, and any $0^m1^\ell0^\omega$ is a name for $\ell < k$. 

\begin{definition}
\label{def:classifier}
A \emph{classifier} is a (computable) procedure that takes some $x \in \mathbf{X}$ as input, and either outputs a colour $j \in \mathbf{k}$, or diverges (which is seen as outputting $\bot$). The collection of classifiers is the space $\mathcal{C}(\mathbf{X},\mathbf{k}_\bot)$.
\end{definition}

At first, we do not need to put restrictions on the domain, and can work just with some represented space $\mathbf{X}$. We will discuss reasons for restrictions later on, in particular in Section \ref{sec:whatspaces}. 

\begin{example}
Let us consider the classifier we would obtain from Support Vector Machine \cite{hearst1998support}. Here, the relevant space $\mathbf{X}$ will be $\mathbb{R}^n$ for some $n \in \mathbb{N}$. The classifier is described by a hyperplane $P$ splitting $\mathbb{R}^n$ into two connected components $C_0$ and $C_1$. We have two colours (corresponding to the two connected components), so the classifier is a map $p : \mathbb{R}^n \to \mathbf{2}_\bot$. If $x \in C_i$, then $p(x) = i$. If $x \in P$, then $p(x) = \bot$.

One reason why the \emph{no-answer answer} $\bot$ has to be treated special is that we will never be able to be certain that a numerical input is exactly on the separating hyperplane (essentially as equality on reals is not decidable). If we are willing to compute with sufficiently high precision, then for any input not on the hyperplane we can eventually figure out on which side it is.

This idealisation is not strictly necessary to make Definition \ref{def:classifier} applicable. One could apply floating point computations and keep track of the errors. In case the errors become too large to classify an input, the response is $\bot.$
This leads to an approximate classifier $\tilde{p} : \mathbb{R}^n \to \mathbf{2}_\bot$, where $\tilde{p}(x) = i$ signifies that with the available accuracy we can confirm that $x \in C_i$, and $\tilde{p}(x) = \bot$ stands for not being able to reach a conclusive answer at the given accuracy level. 
\end{example}

\begin{example}
Neural networks classifier compute a class score for every colour, which when these class scores are normalised, share similar properties as a probability distribution. This translates into our framework by fixing a threshold $p \geq 0.5$, and then assigning a particular colour to an input iff its class score exceeds the threshold $p$. If no colour has a sufficiently high score, the output is $\bot$. As long as the function computing the class scores is computable, so is the classifier we obtain in this fashion. If our class scores can use arbitrary real numbers, we cannot assign a colour for the inputs leading to the exact threshold.

We can, should we choose to do, however add a special colour for those inputs where all scores are strictly below the threshold. It is questionable how desirable assigning colour for the undecided cases is. For example, consider the binary classification task of predicting if an image depicts a cat or a dog. We start with a picture of a cat, and apply several iterations of destructive random image manipulations (such as rotations, sheers, and changing of individual pixel values). Should $\bot$ be a definite class, there would be a computable exact point at which the image ceases to depict a cat \cite{pmlr-v80-athalye18b}.
\end{example}

\paragraph*{Basic verification questions}
What are the verification-questions we could ask about a given classifier $f$ (that is, without considering the learning process that led us to it)? Obviously, we may want to confirm individual requirements of the form $f(x) = n$, but not much sophistication is needed here. It becomes more interesting once we deal with assertions regarding the behaviour of the classifier on an entire set or `region'. While this will generally involve a quantifier over infinite sets of inputs, we can do the following:

\begin{proposition}
\label{prop:basic}
The following maps are computable:
\begin{enumerate}
\item $\mathrm{existsValue} : \mathbf{k} \times \mathcal{V}(\mathbf{X}) \times \mathcal{C}(\mathbf{X},\mathbf{k}_\bot) \to \mathbb{S}$, which answers $\operatorname{true}$ on input $(n,A,f)$ iff $\exists x \in A \ f(x) = n$.
\item $\mathrm{forallValue} : \mathbf{k} \times \mathcal{K}(\mathbf{X}) \times \mathcal{C}(\mathbf{X},\mathbf{k}_\bot) \to \mathbb{S}$, which answers $\operatorname{true}$ on input $(n,A,f)$ iff $\forall x \in A \ f(x) = n$.
\item $\mathrm{fixedValue} : \mathbf{k} \times (\mathcal{V}\wedge\mathcal{K})(\mathbf{X}) \times \mathcal{C}(\mathbf{X},\mathbf{k}_\bot) \to \mathbf{2}_\bot$, which on input $(n,A,f)$ answers $1$ iff $\forall x \in A \ f(x) = n$, and answer $0$ iff $\exists x \in A \ f(x) \in \mathbf{k} \setminus \{n\}$, and $\bot$ otherwise.
\item $\mathrm{constantValue} : (\mathcal{V}\wedge\mathcal{K})(\mathbf{X}) \times \mathcal{C}(\mathbf{X},\mathbf{k}_\bot) \to \mathbf{2}_\bot$, which on input $(A,f)$ answers $1$ iff there is some $n \in \mathbf{k}$ such that $\mathrm{fixedValue}(n,A,f)$ answers $1$, and which answers $0$ iff $\mathrm{fixedValue}(n,A,f)$ answers $0$ for all $n \in \mathbf{k}$.
\end{enumerate}
\end{proposition}

\begin{proof}
\begin{enumerate}
    \item Given a colour $k \in \mathbf{k}$ and a classifier $f \in \mathcal{C}(\mathbf{X},\mathbf{k}_{\bot})$, we can compute $f^{-1}(k) \in \mathcal{O}(\mathbf{X})$ due to the discreteness of $\mathbf{k}$. By definition of overtness, given $A \in \mathcal{V}(\mathbf{X})$, we can then semidecide whether $A \cap f^{-1}(k) \neq \emptyset$.
    \item As above, we can compute $f^{-1}(k) \in \mathcal{O}(\mathbf{X})$. By the definition of compactness, given $A \in \mathcal{K}(\mathbf{X})$ we can semidecide whether $A \subseteq f^{-1}(k)$.
    \item By running the algorithms from 1. and 2. in parallel.
    \item Since $\mathbf{k}$ is both compact and overt, we can quantify over it both existentially and universally. Or, more down to earth, we can simply run the algorithm from 3. for all finitely many cases in parallel, and answer once we have received enough information from those runs.
\end{enumerate}
\end{proof}

One useful application of the map $\mathrm{constantValue}$ is using it on some \emph{small} regions that we are interested in. In ML terms it addresses the question if there are adversarial examples for a classifier in the vicinity of $x$. To characterise small regions, we would have available a metric, and then wish to use closed balls $\overline{B}(x,r)$ as inputs to $\mathrm{constantValue}$.

To be able to do this, we need to obtain closed balls $\overline{B}(x,r)$ as elements of $(\mathcal{V}\wedge\mathcal{K})(\mathbf{X})$. Being able to obtain $\overline{B}(x,r) \in \mathcal{K}(\mathbf{X})$ for sufficiently small $r$, amounts to effective local compactness of $\mathbf{X}$. We generally get $\mathrm{cl} B(x,r)$, the closure of the open ball, as elements of $\mathcal{V}(\mathbf{X})$. For all but countably many radii $r$ we have that $\overline{B}(x,r) = \mathrm{cl} B(x,r)$, and we can effectively compute suitable radii within any interval
\cite{pauly-locallycompact}.

\begin{theorem}
\label{theo:adversarialcomputable}
Let $\mathbf{X}$ be an effectively locally compact computable metric space with metric $d$. There is a computable multi-valued function producing on input $x \in \mathbf{X}$ some real $E^x > 0$ as well as an operation $\mathrm{locallyConstant}_x : (0,E^x) \times \mathcal{C}(\mathbf{X},\mathbf{k}_\bot) \to \mathbf{2}_\bot$, where $\mathrm{locallyConstant}_x(r,f) = 1$ iff $\forall y \in \overline{B}(x,r) \ f(x) = f(y) \neq \bot$, and $\mathrm{locallyConstant}_x(r,f) = 0$ iff $\exists y_0,y_1 \in B(x,r) \ \bot \neq f(y_0) \neq f(y_1) \neq \bot$.

If $(\mathbf{X},d)$ is such that every closed ball is compact, we can ignore the $E^x$ and have $\mathrm{locallyConstant} : \mathbf{X} \times \mathbb{R}^+ \times \mathcal{C}(\mathbf{X},\mathbf{k}_\bot) \to \mathbf{2}_\bot$ instead, with $\mathrm{locallyConstant}(x,r,f) = \mathrm{locallyConstant}_x(r,f)$.
\end{theorem}

\begin{proof}
By \cite[Proposition 12]{pauly-locallycompact}, given a point $x$ in an effectively locally compact computable metric space we can compute a radius $E^x$ and the corresponding closed ball $\overline{B}(x,E^x)$ as a compact set. Subsequently, we can obtain every closed ball as a compact set given a radius less than $E^x$. If every closed ball is compact, we can even obtain them computably as elements of $\mathcal{K}(\mathbf{X})$ by \cite[Proposition 10]{pauly-locallycompact}.

In either case, we have $\overline{B}(x,r)$ available to us as a compact set, and can thus correctly answer $1$ (if this is the case) by running $\mathrm{forallValue}$ from Proposition \ref{prop:basic} for all finitely many possible colours and $\overline{B}(x,r)$.

Without any constraints on the computable metric space $\mathbf{X}$ we can obtain $\mathrm{cl} B(x,r) \in \mathcal{V}(\mathbf{X})$, and we note that if there are two points assigned to distinct colours in $\mathrm{cl} B(x,r)$, there are already such points in $B(x,r)$. Hence, running $\mathrm{existsValue}$ from Proposition \ref{prop:basic} on all finitely many possible colours and $\mathrm{cl} B(x,r)$ lets us correctly answer $0$, if this is the case were are in.
\end{proof}

In the parlance of ML, the map $\mathrm{locallyConstant}_x$ states whether or not their are \emph{adversarial examples} for the classifier $f$ in the vicinity of $x$. If there are no adversarial examples, but some unclassified points (i.e.~where $f$ returns the value $\bot$), then $\mathrm{locallyConstant}_x$ cannot give an answer (because in general, we can never rule out that a classifier will give an answer later on).

\subsection{Adversarial examples}

An adversarial example is the result of a small change or perturbation to the original input that results in a change of classification made by the DNN. I.e.~given the classifier $f$ and an input $x$, an adversarial example is $f(x) \neq f(x + r)$ for $|| r || \leq \epsilon$ and $\epsilon > 0$. It was assumed that many DNNs followed many existing kernel methods in computer vision, where even with small perturbations, the output would still be given the correct class with a high probability. However, \cite{szegedy2013intriguing} demonstrates that this assumption does not hold, even for state-of-the-art image classifiers. They demonstrate the use of a box-constrained optimisation (L-BFGS) method to locate small low-probability `pockets' around the original image that produce miss-classification of a target class. Moreover, this work shows the susceptibility to adversarial examples are not simply an artefact of the training methods (as a model trained on a different subset of the data would produce the same miss-classification), but rather propose there is a deeper and not fully understood meaning behind the effect.

While \cite{szegedy2013intriguing} proposed to use the adversarial examples as part of the training set, the optimisation process was computationally expensive. Later work by \cite{Goodfellow2015} explores a possible reason for adversarial examples existing as well as creating a faster method of locating them. Their Fast Gradient Sign Method (FGSM) automatically generates adversarial examples by adding the sign of the gradient w.r.t. the cost function to the original image. By doing so, it is purposefully pushing the pixel value of the image in the direction that increases the loss of the prediction, and thus possibly creating a miss-classification. This approach allows for adversarial examples to be included in the set of training data, and thus aims to improve the robustness against adversarial examples at test time. 

A method by \cite{Papernot2016a} computes the Jacobian matrix of the input image to find the most significant or salient pixels of the image in which to perturb. The output of their method, Jacobian-based Saliency Map Attack or JSMA, differs from \cite{Goodfellow2015}. Where as in FGSM the perturbations are imperceptible changes to the entire image, JSMA makes a large and noticeable change to a small region, while the rest of the image remains untouched. While showing their method can outperform many other adversarial attacks in various scenarios, its reliance on computing the Jacobian, a potentially slow computation, restricts its usability in many applications. The C\&W attack designed by \cite{Carlini2017} is very effective against many adversarial defences. Their work improves upon the objective function of \cite{szegedy2013intriguing}, to improve the optimisation of the method. In this work they proposed many alternatives for objective function candidates as well as distance metrics.

Despite many methods being proposed to easily fool DNNs, very few quantify and compute the robustness w.r.t. adversarial examples. The \emph{DeepFool} algorithm proposed by \cite{moosavi2016deepfool} aims to compute adversarial examples and compare the robustness of various classifiers to particular types of perturbations. DeepFool iteratively perturbs the original image in the direction of the closest hyperplane and thus creating a miss-classification with a minimal Euclidean distance to the original image as well in a potentially quicker time as compared with \cite{Goodfellow2015}.

All of these methods demonstrate the key principle: image-based DNNs can be easily fooled with precise pixel manipulation. When combined with the realistic scenario of self-driving vehicles, safety concerns arise. \cite{Wicker2018a} use a Gaussian mixture model to identify keypoints in images that describe the saliency map of DNN classifiers. Modifying these keypoints may then change the classification label made by said DNN. They explore their approach on `traffic light challenges' (publicly available dashboard images of traffic lights with red/green annotations). In this challenge they find modifying a single pixel is enough to change neural network classification.

\subsection{Choosing the distance parameter}
  
If we want to use the operation from Theorem \ref{theo:adversarialcomputable} to test for the existence of adversarial examples, we need to know how to choose the parameter $r$ indicating how small perturbations need to be to potentially count as adversarial examples. For example, assume that we want to use our classifier to classify measurement results with some measurement errors. If our measurements are only precise up to $\varepsilon$, then having an adversarial example for $r = \varepsilon$ tells us that we cannot trust the answers from our classifier. We could just as well have received a slightly different answer from our measurement procedure, on which the classifier would have answered differently.

We could also use domain knowledge to select the radius $r$ \cite{morgan2021adaptive}. For example, in an image classification task we could assert a priori that changing a few pixels only can never turn a picture of an elephant into a picture of a car. If we use Hamming distance as metric on the pictures, stating what we mean with \emph{a few pixels} gives us the value $r$ such that any adversarial example demonstrates a fault in the classifier. Another example by \cite{Ruan2018} finds the upper and lower bounds of the input space via an optimisation procedure, following that DNNs are Lipschitz continuous functions and all values between these bounds are reachable.

Here we present a different route: rather than starting with a preconceived notion of small disturbance and then looking for adversarial examples, we can instead compute how much we would need to disturb a given point. In our results, we use the spaces $\mathbb{R}_<$ and $\mathbb{R}_>$, where reals are represented as limits of increasing respectively decreasing rational sequences; as well as the variant $\overline{\mathbb{R}}_<$ where $+\infty$ is included.

\begin{proposition}
\label{prop:radiuslower}
Let $\mathbf{X}$ be an effectively locally compact computable metric space with metric $d$ such that all closed balls are compact. Then the map $$(x,f) \mapsto \sup \{ r \in \mathbb{R} \mid \exists i \in \mathbf{k} \ \forall y \in \overline{B}(x,r) \quad f(y) = i\} : \mathbf{X} \times \mathcal{C}(\mathbf{X},\mathbf{k}_\bot) \to \overline{\mathbb{R}}_<$$ is computable.
\end{proposition}

\begin{proof}
As $\mathbf{k}$ is overt and $\overline{B}(x,r)$ is uniformly compact by assumption (via \cite[Proposition 10]{pauly-locallycompact}), it follows that $$ \exists i \in \mathbf{k} \ \forall y \in \overline{B}(x,r) \quad f(y) = i$$ is semidecidable in the parameters $x$, $f$ and $r$, thus $$\{ r \in \mathbb{R} \mid \exists i \in \mathbf{k} \ \forall y \in \overline{B}(x,r) \quad f(y) = i\} \in \mathcal{O}(\mathbb{R})$$ is an open set computable from $x$ and $f$. The supremum of a set $U \in \mathcal{O}(\mathbb{R})$ is computable as an element of $\overline{R}_<$ as shown in \cite[Section 10]{pauly-synthetic}.
\end{proof}

\begin{proposition}
\label{prop:radiusupper}
Let $(\mathbf{X},d)$ be a computable metric space. Then the map $$(x,f) \mapsto \inf \{r \in \mathbb{R}^{\geq 0} \mid \exists y \in B(x,r) \ \bot \neq f(x) \neq f(y) \neq \bot\} : \mathbf{X} \times \mathcal{C}(\mathbf{X},\mathbf{k}_\bot) \to \mathbb{R}_>$$ is computable.
\end{proposition}

\begin{proof}
It makes no difference whether we use $\exists y \in B(x,r)$ or $\exists y \in \mathrm{cl} B(x,r)$ here. In a computable metric space $\mathbf{X}$, we can compute $\mathrm{cl} B(x,r) \in \mathcal{V}(\mathbf{X})$ from $x$ and $r$. In addition, we observe that $\bot \neq f(x) \neq f(y) \neq \bot$ is semidecidable in $x,y$ and $f$. Alltogether, this gives us access to $$\{r \in \mathbb{R}^{\geq 0} \mid \exists y \in B(x,r) \ \bot \neq f(x) \neq f(y) \neq \bot\} \in \mathcal{O}(\mathbb{R}^{\geq 0})\},$$ and we can then compute the infimum in $\mathbb{R}_>$. Since we consider the infimum in the space $\mathbb{R}^{\geq 0}$, it is at least $0$. 
\end{proof}

\begin{corollary}
Let $\mathbf{X}$ be an effectively locally compact computable metric space with metric $d$ such that all closed balls are compact. The map $\mathrm{OptimalRadius} : \subseteq \mathbf{X} \times \mathcal{C}(\mathbf{X},\mathbf{k}_\bot) \to \mathbb{R}$ defined by $(x,f) \in \dom(\mathrm{OptimalRadius})$ iff $f(x) \neq \bot$, $\exists y \ \bot \neq f(y) \neq f(x)$ and $\forall r, \varepsilon > 0 \ \exists z \in B(x,r+\varepsilon) \setminus B(x,r) \ f(z) \neq \bot$; and by $$\mathrm{OptimalRadius}(x,r) = \sup \{ r \in \mathbb{R} \mid \exists i \in \mathbf{k} \ \forall y \in \overline{B}(x,r) \quad f(y) = i\} = \inf \{r \mid \exists y \in B(x,r) \ \bot \neq f(x) \neq f(y) \neq \bot\}$$ is computable.
\end{corollary}

\section{Learners and their robustness}
\label{sec:robust_learners}

Let us now consider the process of training the classifier. To keep matters simple, we will not adopt a dynamic view, but rather model this as a one-step process. We also only consider supervised learning. This leads us to the conception of a learner as a map from finite sequences of labelled points to classifiers:

\begin{definition}
A \emph{learner} is a (computable) procedure that takes as an input a tuple $((x_0,n_0),\ldots,(x_\ell,n_\ell)) \in (\mathbf{X} \times \mathbf{k})^*$ and outputs a classifier $f \in \mathcal{C}(\mathbf{X},\mathbf{k}_\bot)$. The collection of all learners is the space $\mathcal{C}((\mathbf{X} \times \mathbf{k})^*,\mathcal{C}(\mathbf{X},\mathbf{k}_\bot))$.
\end{definition}

\paragraph*{Faithful reproduction of the training labels}
We do not prescribe any particular relation between the training data and the behaviour of the resulting classifier. It could seem reasonable to ask that a learner $L$ faithfully reproduces the training data, i.e.~satisfies $L((x_i,n_i)_{i \leq \ell})(x_m) = n_m$, but such a criterion is impossible to satisfy, as our notion of training data does not rule out having multiple occurrences of the same sample point with different labels. It would also not match applications, as it often is desirable that a model can disregard parts of its training data as being plausibly faulty.

\begin{proposition}
\label{prop:doesdeviate}
Let $\mathbf{X}$ be computably overt and computably Hausdorff. The operation \[\mathrm{doesDeviate} : \mathcal{C}((\mathbf{X} \times \mathbf{k})^*,\mathcal{C}(\mathbf{X},\mathbf{k}_\bot)) \to \mathbb{S}\] returning $\operatorname{true}$ on input $L$ iff there is some input $(x_i,n_i)_{i \leq \ell} \in (\mathbf{X} \times \mathbf{k})^*$ with $x_i \neq x_j$ for $i \neq j$, and some $m \leq \ell$ such that $L((x_i,n_i)_{i \leq \ell})(x_m) \in \mathbf{k} \setminus \{n_m\}$ is computable.
\end{proposition}
\begin{proof}
If $\mathbf{X}$ is overt, then so is $(\mathbf{X} \times \mathbf{k})^*$. Since the condition for answering  $\operatorname{true}$ is defined by existentially quantifying a semidecidable condition over an overt set, it is semidecidable itself.
\end{proof}

\paragraph*{Robustness under additional training data}
Generally, our goal will not be so much to algorithmically verify properties of learners for arbitrary training data, but rather be interested in the behaviour of the learner on the given training data and hypothetical small additions to it. One question here would be to ask how robust a classifier is under small additions to the training data. A basic version of this would be: 

\begin{proposition}
\label{prop:robustpoint}
Let $\mathbf{X}$ be computably compact and computably overt. The map \[\operatorname{robustPoint} : \mathbf{X} \times (\mathbf{X} \times \mathbf{k})^* \times \mathcal{C}((\mathbf{X} \times \mathbf{k})^*,\mathcal{C}(\mathbf{X},\mathbf{k}_\bot)) \to \mathbf{2}_\bot\] answering $1$ on input $x$, $(x_i,n_i)_{i \leq \ell}$ and $L$ iff \[\forall x_{\ell+1} \in \mathbf{X} \ \forall n_{\ell+1} \in \mathbf{k} \quad L((x_i,n_i)_{i \leq \ell})(x) = L((x_i,n_i)_{i \leq \ell + 1})(x) \in \mathbf{k}\] and answering $0$ iff \[\exists x_{\ell+1} \in \mathbf{X} \ \exists n_{\ell+1} \in \mathbf{k} \quad \bot \neq L((x_i,n_i)_{i \leq \ell})(x) \neq L((x_i,n_i)_{i \leq \ell + 1})(x) \neq \bot\]  is computable.
\end{proposition}
\begin{proof}
Since $\mathbf{X}$ and $\mathbf{k}$ are both compact and overt, both universal and existential quantification preserves semidecidable predicates. Since $\mathbf{k}$ is discrete and Hausdorff, the stem of our predicate definitions for answering $0$ and $1$ are both semidecidable.
\end{proof}

The map $\operatorname{robustPoint}$ can confirm for us that a particular classification would remain, even if we add another point to the sample; or that a labeling can change upon extending the training data a little bit. The very same reasoning would also work if we would consider adding two, three or $j$ additional samples. We can also lift this to compact and overt regions, and then ask whether either all values will stay unchanged, or whether there is one that does. As usual, if the point (or any point in the region) is assigned $\bot$, $\operatorname{robustPoint}$ will output $\bot$, too.

\paragraph*{Sparsity of training data}
Allowing arbitrary additional training data as above might not be too suitable -- for example, if we add the relevant query point together with another label to the training data, it would not be particularly surprising if the new classifier follows the new data. If we bring in a metric structure, we can exclude new training data which is too close to the given point.

\begin{definition}
\label{def:sparse}
Fix a learner $L : (\mathbf{X} \times \mathbf{k})^* \to \mathcal{C}(\mathbf{X},\mathbf{k}_\bot)$, some $N \in \mathbb{N}$ and $\varepsilon > 0$. We say that $(x_i,n_i)_{i \leq \ell}$ is \emph{sparse} at $x \in \mathbf{X}$, if there are $(y_i,m_i)_{i \leq j}$ and $(y'_i,m'_i)_{i \leq j'}$ such that $\ell + N \geq j,j' \geq \ell$, $y_i = y'_i = x_i$ and $m_i = m'_i = n_i$ for $i \leq \ell$, and $d(y_i,x),d(y'_i,x) > \varepsilon$ for $i > \ell$ satisfying $\bot \neq L((y_i,m_i)_{i \leq j})(x) \neq L((y'_i,m'_i)_{i \leq j'})(x) \neq \bot$. 

We say that $(x_i,n_i)_{i \leq \ell}$ is \emph{dense} at $x \in \mathbf{X}$ if for all $(y_i,m_i)_{i \leq j}$ and $(y'_i,m'_i)_{i \leq j'}$ such that $\ell + N \geq j,j' \geq \ell$, $y_i = y'_i = x_i$ and $m_i = m'_i = n_i$ for $i \leq \ell$, and $d(y_i,x),d(y'_i,x) \geq \varepsilon$ for $i > \ell$  it holds that $L((y_i,m_i)_{i \leq j})(x) = L((y'_i,m'_i)_{i \leq j'})(x) \neq \bot$.
\end{definition}

To put it in words: Training data is dense at a point whose label it determines, even if we add up to $N$ additional points to the training data, which have to be at least $\varepsilon$ away from that point. Conversely, at a sparse point we can achieve different labels by such an augmentation of the training data. If we have chosen the parameters $N$ and $\varepsilon$ well, then we can conclude that based on the training data we can make reasonable assertions about the dense query points, but unless we have some additional external knowledge of the true distribution of labels, we cannot draw reliable conclusion about the sparse query points. We concede that it would make sense to include points under \emph{sparse} where the classifiers will always output $\bot$ even if we enhance the training data, but this would destroy any hope of nice algorithmic properties.

\begin{theorem}
\label{theo:sparsity}
Let $\mathbf{X}$ be a computably compact computable metric space. The operation \[\operatorname{SprsOrDns} : \mathcal{C}((\mathbf{X} \times \mathbf{k})^*,\mathcal{C}(\mathbf{X},\mathbf{k}_\bot)) \times \mathbb{N} \times \mathbb{R}_+ \times (\mathbf{X} \times \mathbf{k})^* \times \mathbf{X} \to \mathbf{2}_\bot\] answering $0$ on input $L,N,\varepsilon$, $(x_i,n_i)_{i \leq \ell}$ and $x$ iff $(x_i,n_i)_{i \leq \ell}$ is sparse at $x$, and answers $1$ if $(x_i,n_i)_{i \leq \ell}$ is dense at $x$ is computable.
\end{theorem}

\begin{proof}
Essentially, the claim is that both being sparse and being dense is a semidecidable property. We show this by examining Definition \ref{def:sparse} and in particular its quantification structure. 

We note that $\bot \neq L((y_i,m_i)_{i \leq j})(x) \neq L((y'_i,m'_i)_{i \leq j'})(x) \neq \bot$ is a semidecidable property. We are then adding existential quantifications over $\operatorname{cl} \{y \in \mathbf{X} \mid d(x,y) > \varepsilon\}$, and this set is always available to us as an overt set when working in a computable metric space. Thus, being sparse is semidecidable.

For density, note that $L((y_i,m_i)_{i \leq j})(x) = L((y'_i,m'_i)_{i \leq j'})(x) \neq \bot$ is again semidecidable. This time we add universal quantification over $\{y \in \mathbf{X} \mid d(x,y) \geq \varepsilon\}$. This set is available to us as a compact set thanks to the demand that $\mathbf{X}$ be a computably compact computable metric space.

\end{proof}

\paragraph*{Sparsity in the ML literature}

In the context of this study, sparsity refers to the \emph{measurement} of the input representation, and is not the representation itself (i.e.~sparse matrix forms). Indeed, we may evaluate the \emph{sparsity} of sampling from any underlying manifold that may have further implications on how ML algorithms learn and create classifications in these particular regions of space. In sparse regions of a manifold, there will be less direct information for the ML algorithm to learn from. Regions such as these then may harbour many more adversarial examples \cite{Lindenbaum2018GeometryGeneration}. Such methods of measurement may be useful for the creation of adversarial examples (or at least the search for adversarial examples) by manipulating the region size of our closed ball $\overline{B}(x, r)$ relative to the local information of sparsity of sampling of the manifold space. 

One typical measurement for sparsity of sampling within data with kernel density estimation (KDE), commonly with a Gaussian kernel $\text{exp} \left( \frac{|| \mathbf{x} - \mathbf{x}' ||^2}{2\sigma^2} \right)\,$  where $\sigma$ controls the width of the kernel. For small values of $\sigma$ the similarity matrix forms an identity matrix, or a matrix of all-ones for large values. Thus choosing an optimal bandwidth parameter is problem specific, and may require tuning to appropriately quantify the sparsity of the manifold space. However, some research has been conducted to alleviate this issue. Two methods are described within \cite{Lindenbaum2018} for the setting of bandwidth parameters in KDE that are locally adaptive to the regions, thereby optimally characterising the sparsity measurement with respect to each sample of the manifold space.

\section{Musings on the class of spaces to consider}
\label{sec:musings}

\label{sec:whatspaces}
In this section we focus on what demands we should make on the space $\mathbf{X}$. While we have seen that many operations can be defined and shown to be computable in a rather general setting, we will make the case here that effectively locally-compact computable metric spaces form a particular convenient setting. They have enough structure to make (almost) everything we have considered make sense and computable, are sufficiently general to contain conceivable settings from applications.

\paragraph*{Impact of names}
Usually, points $x \in \mathbf{X}$ will have multiple names $p \in \delta_\mathbf{X}^{-1}(\{x\})$. For example, Brattka and Hertling have shown that no admissible representation of the real numbers makes all reals have only finitely many names \cite{hertling7}. As a consequence, an algorithm can fail to compute an extensional function by giving different answers when reading different names of the input.

A desideratum would be that we could modify a given algorithm such that the modified algorithm gives a certain answer on an input only if the original algorithm gives that very same answer on all other names of the same input. Being able to do this means being able to universally quantify over names, which in turn means that we need an effectively fiber-compact representation (which is defined as letting us compute the compact set of names of a point from that point). It was shown in \cite{pauly-kihara-arxiv} that a space has an effectively fiber-compact computably admissible representation iff it is computably metrizable.

We could be less cautious, and instead decide to alter our algorithms such that producing a colour when reading some name of a point should lead to all names being mapped to that colour. This runs into the issue that there is no computable way to handle tie-breaks between different names mapping to different colours. We can handle this by adjoining to $\mathbf{k}_\bot$ a further element $\top$, where any sequence ending in $1^\omega$ denotes $\top$. This allows a computation to change from any colour to $\top$. While $\bot$ denotes \emph{no information}, $\top$ would denote \emph{contradictory information}. We still need to be able to existentially quantify over names, meaning that we need effectively fiber-overt representations. Having an effectively fiber-overt representation is equivalent to being effectively countably-based.

\paragraph*{Complexity theory}
While we have consciously avoided complexity issues in the present paper, we cannot deny that efficiency of algorithms will be a crucial issue for practical relevance. There is a complexity theory for computable analysis (see in particular Kawamura and Cook \cite{kawamura}, or for a generalisation \cite{eike2}), however, here the run-times of algorithms are no longer described by functions $\tau : \mathbb{N} \to \mathbb{N}$, but rather by second-order objects (e.g.~functions $\Upsilon : \mathbb{N}^\mathbb{N} \times \mathbb{N} \to \mathbb{N}$). However, this price is not due immediately when considering algorithms on non-discrete domains. Schr\"oder \cite{schroder6} has characterized the class of spaces admitting a complexity theory such that runtimes are still functions $\tau : \mathbb{N} \to \mathbb{N}$, these are the coPolish spaces\footnote{The term \emph{coPolish} was coined only after the publication of \cite{schroder6}. Its earliest occurrence in the publication record seems to be \cite{paulydebrecht4}.}. A separable metric space is locally compact iff it is coPolish, and a countably-based coPolish space is already metrizable. As such, we see that with the backdrop of our other desiderata, being locally compact and admitting a nice complexity theory are intricately linked.

\paragraph*{Hausdorffness}
Requiring spaces to be Hausdorff is certainly a popular requirement, even in the absence of necessity. In theoretical computer science we do, however, encounter many non-Hausdorff topological spaces (e.g.~\cite{goubault}). Most of the spaces we have referred to in this work are non-Hausdorff, such as the spaces $\mathbf{k}_\bot$. We do desire the space $\mathbf{X}$ to be Hausdorff, though. It is worth pointing out that this requirement can  be formulated in terms of discriminating classifiers:

\begin{proposition}
\label{prop:separatingfunctions}
The following are equivalent for a  countably-based  represented space $\mathbf{X}$:
\begin{enumerate}
\item $\mathbf{X}$ is Hausdorff.
\item For any $x, y \in \mathbf{X}$ with $x \neq y$ there exists a computable classifier $f_{xy} : \mathbf{X} \to \mathbf{2}_\bot$ with $f(x) = 0$ and $f(y) = 1$.
\end{enumerate}
\end{proposition}
\begin{proof}
If $\mathbf{X}$ is countably-based and Hausdorff, then distinct points are separated by disjoint basic open sets. If $U$ and $V$ are disjoint basic open sets, then $f_{UV} : \mathbf{X} \to \mathbf{2}_\bot$ with $f^{-1}(0) = U$ and $f^{-1}(1) = V$ is well-defined and computable.

A separating classifier $f_{xy} : \mathbf{X} \to \mathbf{2}_\bot$ immediately yields disjoint open sets separating $x$ and $y$, and thus witnesses the Hausdorff condition.
\end{proof}

\paragraph*{Schr\"oder's effective metrization theorem}
An effectively locally-compact, effectively countably-based computably Hausdorff is computably metrizable by Schr\"order's effective metrization theorem \cite{grubba3}. If we are willing to work relative to an oracle, a locally-compact represented space is already countably based by \cite[Corollary 6.11]{escardo7}. Thus, if we are asking for local compactness anyway to be able to verify that a classifier is locally constant (i.e.~has no adversarial examples in a neighbourhood of a given point), we can just as well as for the existence of a metric and make available to us a clear definition of adversarial examples. Of course, a metrization theorem does not tell us which metric to use. 

\paragraph*{Compactness}
Some of our results (Proposition~\ref{prop:robustpoint}, Theorem~\ref{theo:sparsity}) even require compactness, not just local compactness. While we could demand that the spaces we handle are compact, this could also be solved by restricting the universal quantification to a suitable compact neighbourhood of all the points concerned. We have not included this in our formal statements to avoid near-unreadable complexity.

\section{Future work}

The results in this paper assert that under certain conditions on the space, certain maps are computable. Dropping these conditions, or considering maps providing more information instead, generally will lead to non-computability. We leave the provision of counterexamples, as well as potentially a classification of \emph{how}\footnote{This would be answered using the formalism of Weihrauch degrees\cite{pauly-handbook}.} non-computable these maps are to future work. The notion of a \emph{maximal partial algorithm} recently proposed by Neumann \cite{eike-neumann2} also seems a promising approach to prove optimality of our results.

There is a trade-off between the robustness of a classifier and its `accuracy'. It seems possible to develop a computable quantitative notion of robustness on top of our function locallyConstant, which could then be used as part of the training process in a learner. This could be a next step to adversarial robustness \cite{Carlini2017,Goodfellow-et-al-2016}.

Rather than just asking questions about particular given classifiers or learners, we could start with a preconception regarding what classifier we would want to obtain for given training data. Natural algorithmic questions then are whether there is a learner in the first place that is guaranteed to meet our criteria for the classifiers, and whether we can compute such a learner from the criteria.






\bibliography{references}

\end{document}